\documentclass{article}

\usepackage[nonatbib, preprint]{neurips_2021}
\usepackage[utf8]{inputenc} %
\usepackage[T1]{fontenc}    %
\usepackage{hyperref}       %
\usepackage{url}            %
\usepackage{booktabs}       %
\usepackage{amsfonts}       %
\usepackage{nicefrac}       %
\usepackage{microtype}      %
\usepackage[english]{babel}
\usepackage{graphicx}
\usepackage{biblatex}
\usepackage{csquotes}
\usepackage{amsmath}
\usepackage{bbm}
\usepackage{qtree}
\usepackage{algpseudocode, algorithm}
\algnewcommand\EOLCOMMENT{\item[\algorithmicinput]}

\usepackage[table,xcdraw]{xcolor}
\usepackage[makeroom]{cancel}
\usepackage{caption}
\usepackage{subcaption}
\usepackage{adjustbox}

\usepackage{fancyhdr}
\usepackage{color}
\usepackage{url}
\usepackage{multirow}
\usepackage{graphicx} %
\usepackage{times}
\usepackage{latexsym}
\usepackage{multicol}
\usepackage{multirow}

\newcommand{\qnote}[1]{}

\usepackage{tabularx} %
\usepackage{amssymb}%
\usepackage{amsmath}
\usepackage{mathtools}
\usepackage{pifont}%

\usepackage{ragged2e}

\usepackage{etoolbox}
\makeatletter
\preto{\@tabular}{\parskip=3pt}
\makeatother

\usepackage{enumitem}
\setlist[itemize]{leftmargin=*}

\usepackage{graphicx}

\title{Towards Automatic Actor-Critic Solutions to Continuous Control}

\author{
  Jake Grigsby \\
  University of Virginia \\
  \texttt{jcg6dn@virginia.edu} \\
  \And
  Jin Yong Yoo \\
  University of Virginia \\
  \texttt{jy2ma@virginia.edu} \\
  \And
  Yanjun Qi \\
  University of Virginia \\
  \texttt{yanjun@virginia.edu} \\
}

\hypersetup{
    colorlinks=true,
    linkcolor=blue,
    filecolor=magenta,      
    urlcolor=cyan,
}
\begin{document}
\maketitle

\begin{abstract}
\begin{quote}
Model-free off-policy actor-critic methods are an efficient solution to complex continuous control tasks. However, these algorithms rely on a number of design tricks and hyperparameters, making their application to new domains difficult and computationally expensive. This paper creates an evolutionary approach that automatically tunes these design decisions and eliminates the RL-specific hyperparameters from the Soft Actor-Critic algorithm. Our design is sample efficient and provides practical advantages over baseline approaches, including improved exploration, generalization over multiple control frequencies, and a robust ensemble of high-performance policies. Empirically, we show that our agent outperforms well-tuned hyperparameter settings in popular benchmarks from the DeepMind Control Suite. We then apply it to less common control tasks outside of simulated robotics to find high-performance solutions with minimal compute and research effort.
\end{quote}
\end{abstract}

\section{Introduction}
Deep Reinforcement Learning (RL) has had great success in many diverse and challenging domains, from robotics \cite{kalashnikov2018qtopt} to the game of Go \cite{silver2017mastering} and autonomous balloon navigation \cite{bellemare_autonomous_2020}. However, day-to-day progress in the field is measured in a limited number of benchmark tasks and tends to be dominated by a small group of algorithms. The model-free off-policy actor-critic literature includes dozens of papers that compare their methods on simulated locomotion tasks that have been popular for half a decade \cite{schulman2015highdimensional} \cite{brockman2016openai}. In that time, the research community has settled on a set of accepted hyperparameters and design heuristics that rarely changes. While this may save time and compute when comparing methods on common benchmarks, it makes approaching a brand new domain a daunting and computationally expensive challenge. Many of the most important hyperparameters in state-of-the-art actor-critic algorithms are unintuitive; even experienced RL practitioners may need to resort to grid search and other expensive hyperparameter optimization techniques.

This paper looks to automate the process of tuning an actor-critic algorithm and creates an out-of-the-box solution to dense-reward continuous control problems. The result is a general algorithm that tunes almost every RL design decision and returns an ensemble of high-performance policies while remaining sample-efficient. First, we compare against baseline approaches in common control benchmarks. Then we evaluate our method's ability to reduce engineering effort in new domains by applying it to complex tasks inspired by real-world industrial challenges and operations research. Our solution, which we call Automatic Actor-Critic (\textbf{AAC}), is easy to implement and leaves several promising directions for future improvement. We have open-sourced an implementation of AAC \href{https://github.com/jakegrigsby/deep_control}{at this GitHub link}.

\section{Background}

\subsection{Model-Free Off-Policy Actor-Critics for Continuous Control}
We assume the reader is familiar with Reinforcement Learning and the general Markov Decision Process (MDP) setting; see Appendix \ref{mdp-notation} for an overview of notation. This paper focuses on solving control tasks where actions are continuous vectors. Deep RL research in continuous control was partly inspired by the success of model-free off-policy actor-critic methods like DDPG \cite{lillicrap2015continuous}. DDPG and later variations iteratively learn a policy $\pi$ and action-value function(s) $Q$, parameterized by a deep neural network actor and critic(s), denoted $\pi_{\theta}$ and $Q_{\phi}$ respectively. Interactions are sampled from the environment and added to a large replay buffer $\mathcal{D}$. Typically we encourage the exploration of new behavior by injecting random noise into the policy or sampling from a stochastic actor. The critic networks are updated by approximate dynamic programming to learn the value of taking action $a$ in a given state $s$. This is accomplished by regression to a temporal difference (TD) target:
\begin{align}
\mathcal{L}_{critic} = \mathop{\mathbb{E}}_{(s, a, r,  s', d) \sim \mathcal{D}}\left[ \big(Q_{\phi}(s, a) - (r + (1 - d)\gamma Q_{\phi'}(s', \Tilde{a}'))\big)^2\right]
\label{critic}
\end{align}

Where $\Tilde{a}' \sim \pi_{\theta}(s')$, and $\phi^\prime$ term refers to target networks, which are discussed in more detail in Section \ref{design_choices}. The continuous action space makes it intractable to recover an optimal policy directly from the $Q$ function. Instead, we train the actor network to maximize the output of the critic network:
\begin{align}
    \mathcal{L}_{actor} &= \mathop{\mathbb{E}}_{s \sim \mathcal{D}}\left[-Q_{\phi}(s, \pi_{\theta}(s))\right]
\label{actor}
\end{align}
State-of-the-art algorithms typically learn an ensemble of critic networks to reduce update bias due to function approximation error \cite{fujimoto2018addressing}. In the case of two critic networks, this is known as the ``clipped-double-Q-trick" - though it can be expanded to an arbitrary number of networks \cite{chen2021randomized}.

The community has generated a vast literature discussing various techniques to improve sample efficiency and performance. Among these are algorithms such as TD3 \cite{fujimoto2018addressing}, SAC \cite{haarnoja2019soft}, SUNRISE \cite{lee2020sunrise}, DisCor \cite{kumar2020discor}, REDQ \cite{chen2021randomized}, and GRAC \cite{shao2020grac} - to name only a few. Model-free off-policy actor-critics retain their popularity because their performance is near state-of-the-art on common benchmarks while being widely available and reproducible. Deep Actor-Critics have wide-ranging applications, but the incremental progress in the field is primarily measured on a small set of benchmark tasks. In the process of inventing, comparing, and re-implementing dozens of alternative approaches on this small task set, the research community has settled on several heuristic design choices that are critical to high performance. The reliance on these settings makes it difficult to apply this family of algorithms to new domains. In the next section, we discuss what we consider to be the most important hyperparameters in off-policy actor-critics and the challenges that come with tuning them. 

\subsection{Design Decisions in Deep Actor-Critic Algorithms}
\label{design_choices}

\textbf{Target networks, Learning Schedules and Replay Ratios: } Target networks (Eq \ref{critic}) prevent the update to $Q_{\phi}(s, a)$ from immediately impacting the value of $Q_{\phi}(s', a')$, which destabilizes learning by altering the value of the $Q$-network's TD target. We can control our targets' rate of change by using a separate network to output $Q_{\phi'}(s', a')$, and updating that network periodically \cite{mnih_human-level_2015} or as a moving average of the online critics' weights with polyak parameter $\tau$ \cite{lillicrap2015continuous}. In either case, this creates an important hyperparameter decision. Updating the target network too quickly or too often will destabilize learning, while updating it too slowly or infrequently leads to an unnecessary drop in sample efficiency. GRAC \cite{shao2020grac} proposes a ``self-regularized" update that preserves stability by explicitly penalizing changes in $Q_{\phi}(s', a')$ - removing the need for target networks and their hyperparameters. However, the penalized update has the side-effect of being much more conservative, requiring several critic gradient updates per training step. The question of precisely how many more updates are needed introduces additional hyperparameters. We need to identify the optimal ratio of 1) environment samples collected, 2) actor gradient updates, 3) critic gradient updates. These values create a \textit{learning schedule} or \textit{replay ratio}. Recent work has found that increasing the replay ratio can boost sample efficiency and performance \cite{fedus2020revisiting} \cite{chen2021randomized}.

\textbf{Action Persistence and Control Frequency: } The control frequency is the rate at which the agent receives states from the environment and is required to provide a new action. In real-world robotics, this might be determined by sensor delay or other natural barriers. The control frequency of simulated research environments is often a relatively arbitrary constant set inside the parameters of the physics simulator (e.g., MuJoCo \cite{6386109}). As control frequency increases, the time between actions decreases, and therefore the consequences of individual actions become difficult to distinguish. More formally, the advantage $Q(s, a) - V(s)$ of an action $a$ approaches zero as the time between states approaches zero \cite{tallec2019making} \cite{metelli2020control}. This presents a challenge to Q-learning methods because the critic value landscape we are maximizing will appear to be mostly flat. A simple way to address this is to enforce a lower control frequency by repeating the agent's past action $k$ times before asking for a new decision. This reduces the control frequency by a factor of $k$ and increases the advantage of optimal actions. The value of $k$ is called the ``action persistence", ``action repeat" or ``frame skip" parameter. Increasing the action persistence can also improve exploration and reduce forward pass computation during deployment \cite{kalyanakrishnan2021analysis} \cite{hessel2019inductive}.

A properly tuned action persistence can have a significant impact on performance \cite{Reda_2020}. However, high values of $k$ can cause the gap between action choices to be too long to adapt to sudden changes in the environment. There have been several proposed ways to tune $k$. One approach is to make the action repeat an aspect of the action itself. In discrete settings, this may involve multiplying the action space to create a new action for each original choice but at several different values of $k$ \cite{DBLP:conf/aaai/Lakshminarayanan17}. In continuous spaces, the persistence can become a new dimension of the action vector \cite{sharma2020learning}. Another approach (TASAC \cite{yu2021tasac}) learns a second policy whose only action choices are to repeat the previous action of the main policy or select a new one. More complicated methods directly estimate the optimal action persistence \cite{metelli2020control} and can handle different control frequencies for each component of the action \cite{NEURIPS2020_216f44e2}.

\textbf{Discount Factor: } The MDP discount factor $\gamma$ controls the time horizon over which the agent is maximizing returns. This value is usually treated as a fixed element of the benchmark and set at $.99$. However, agents are almost always evaluated based on their undiscounted $(\gamma = 1.0)$ returns, which makes $\gamma$ more of an agent-side hyperparameter than an environmental constraint \cite{hessel2019inductive}. Prior work has considered hyperparameter schedules for $\gamma$ that boost performance by regularizing learning \cite{amit2020discount}.

\textbf{Entropy Regularization: } One way to ensure diverse experience collection is to optimize for policy entropy, balanced by an additional hyperparameter $\alpha$. PPO \cite{schulman2017proximal} and A2C/A3C \cite{mnih2016asynchronous} set $\alpha$ to be a small fixed constant. SAC \cite{haarnoja2018soft} uses a MaxEnt-RL framework that makes entropy part of the value function. This increases policy entropy and is thought to have other benefits, including robustness to environmental uncertainty and partially observed reward functions \cite{eysenbach2019maxent}. Entropy regularization also keeps the exploration policy more centered around zero as opposed to the random noise heuristics of TD3 and DDPG, which can cause actions to be repeatedly clipped at $(-1 ,1)$ \cite{wang2020striving}. The follow-up version of SAC \cite{haarnoja2019soft} tunes $\alpha$ to dynamically approach a target entropy level with gradient descent. This target entropy level is denoted $H$ and set to $-|\mathcal{A}|$ - a value that works empirically on benchmark tasks but becomes an unintuitive hyperparameter in new domains.  Meta-SAC \cite{wang2020metasac} tunes the target entropy value with a meta-gradient approach.

\section{Method: Automatic Actor Critics}
\label{method}
The issues above require a number of heuristic solutions that would be expensive to re-tune and re-evaluate on a new domain. We attempt to address this by creating a unified approach that automatically discovers new heuristics for each task and sheds as many hyperparameters as possible. We find inspiration in Population Based Training (PBT) \cite{jaderberg2017population}. In PBT, a population of training runs with different hyperparameters are conducted independently. At regular intervals, the performance of each run is used to generate a more optimal set of hyperparameters according to an evolutionary strategy where the parameters of the highest-performing runs are used to re-initialize the worst-performing setups.  Hyperparameters are randomly perturbed to explore the parameter space. In the off-policy RL context, PBT is quite sample-inefficient because each training run in the population collects a full buffer of samples independently despite being designed to recycle data from a variety of policies. Our first modification is to share environment experience across members of the population. Because the replay buffer we are optimizing over now consists of the experience of many different agents with different parameters, we are diversifying experience collection and recovering the exploration advantages of multi-actor setups like D4PG \cite{barthmaron2018distributed}. 

We initialize a population of $M$ SAC-style actor-critic agents and begin by searching over $\gamma$ and the target entropy coefficient $H$. We make a change of notation from $\gamma$ and $H$. $g$ substitutes $\gamma$, where $\gamma = 1 - exp(g)$; this gives the agent more control over the small differences between discount values approaching $1.0$. $h$ substitutes for the target entropy $H$, where $H = h( -|\mathcal{A}|)$, meaning it is a coefficient for the default SAC heuristic of $H = -|\mathcal{A}|$. 

From there, we add two adjustments to the core agent intended to reduce hyperparameters and find higher-performance policies. We eliminate the need for target networks by utilizing the self-regularizing critic update from GRAC \cite{shao2020grac}. Updates are stabilized by minimizing the impact that changes to $Q(s, a)$ that have on the target value $Q(s', a')$:
\begin{align}
\label{self-regularized}
    \mathcal{L}_{sr} = \mathcal{L}_{critic} + (\cancel{\nabla}Q_{\phi}(s', a') - Q_{\phi}(s', a'))
\end{align}
where $\cancel{\nabla}$ denotes a stop gradient operator. This process can be replicated across each critic network when using the clipped-double-Q trick or another bias-reducing method. This loss function slows critic learning by reducing the impact of each gradient update. The GRAC authors address this by introducing an additional heuristic whereby the critic optimization loop continues until the critic loss is less than some percentage of its initial value on that training step. That percentage is increased throughout training as the critic is more accurate and is less able to reduce its loss function. Experiments in \cite{shao2020grac} demonstrate that this is a sensitive hyperparameter. Our goal is to eliminate sensitive hyperparameters, so we add the number of actor and critic updates per gradient step as separate PBT-tuned parameters. We denote these as $a$ and $c$, respectively. Searching over both $a$ and $c$ creates an adaptive replay ratio schedule that can improve sample efficiency.

The action persistence value $k$ discussed in Sec \ref{design_choices} can have a critical impact on performance. Rather than adding additional action outputs to adjust action repetitions $k$, we experiment with the simpler solution of making $k$ a tunable parameter of the environment and add it to the PBT search. However, adjusting the control frequency of the population's experience over time complicates the use of replay buffer data. We address this by concatenating the current value of $k$ to the state vector of the environment. This allows the actor and critic networks to recognize changes in control frequency and adapt their output accordingly while replaying transitions from the buffer as usual. A side effect of this approach is that it allows the agent to generalize across control frequencies and adapt to changes during deployment. There are some additional details related to how we compensate for changes in $\gamma$ to the reward of frame-skipped transitions. A thorough discussion of this approach to ``persistence-aware actor-critics" is provided in Appendix \ref{sec:persistence_critics} along with additional experiments focused on this idea.

In total, we are now automatically searching over five key hyperparameters $(a, c, k, g, h)$. Each member of the population trains for one evolutionary epoch with its own hyperparameter values. The population is then evaluated in the environment to determine each member's ``fitness" ($f$). We set the fitness of agent $i$, denoted $f_i$, to its mean return with action persistence $k_i$\footnote{It is common for environments to have strict maximum episode lengths that directly influence the final return. In these cases, we compensate for differences in action repetition by dividing the step limit by $k$.}. However, more complex novelty-related bonuses could be incorporated to improve exploration (Sec \ref{limitations}). The highest-fitness members are randomly paired with the lowest-fitness members to transfer and then perturb their hyperparameter values. Network parameters and optimizer states are also transferred.

All that is left to define are the ranges of hyperparameter values that we would like to search. While this may seem like we have traded each parameter for three new ones (the lower bound, upper bound, and random perturbation range), these are intuitive to define in practice. If our range is too broad, the evolutionary algorithm may take more time to find the correct values, but we can be reasonably confident that it will. The only difficult hyperparameters that we have introduced are the frequency of evolutionary updates and the population size. However, both have intuitive runtime/performance tradeoffs - increasing the population size and length of individual training runs makes us more likely to find the correct parameters and more likely to notice the performance gap between them. We can set these meta-parameters in advance based on available time, compute, and problem difficulty.

Pseudocode is provided in Algorithm~\ref{algo} and additional implementation details are discussed in Appendix \ref{implementation}. We will refer to this method as ``Automatic Actor-Critic" (\textbf{AAC}). To summarize, this agent:
\begin{itemize}
    \item Does not use target networks and their associated hyperparameters. We automatically learn the replay ratio and additional critic update schedule.
    \item Dynamically adjusts the action persistence but arrives at a fixed control frequency. As a side effect, it is also capable of adapting to sudden changes in control frequency.
    \item Does not rely on random noise heuristics for exploration. We sample from a high-entropy stochastic policy that is automatically tuned to approach a target entropy level that is also automatically tuned.
    \item Does not treat the discount factor as a fixed environment parameter and can dynamically adjust $\gamma$ to improve evaluation performance.
    \item Improves exploration by sampling experience from a variety of diverse policies.
    \item Has just two important hyperparameters, both of which have intuitive performance/runtime tradeoffs that can be considered in advance.
    \item Returns a population of high-performance solutions that can be ensembled to form a robust final policy.
\end{itemize}

\begin{figure}[tb]
\centering
\begin{minipage}{.8\linewidth}
\begin{algorithm}[H]
\caption{Automatic Actor Critic Training \label{algo}} 
\begin{algorithmic}[1]
    \Require Population size $M$, evolutionary epoch $E$, steps per epoch $T$, min and max values for $a, c, h, k, g$ (denoted as $a_{\min}, a_{\max}, \dots, g_{\min}, g_{\max}$).
    \vspace{2mm}
    \State $\mathcal{D}\leftarrow$ replay buffer initialized with random samples
    \For{$i=1, \dots, M$ in population}
        \State $a_i \sim \mathcal{U}(a_{\min}, a_{\max})$
        \State $c_i \sim \mathcal{U}(c_{\min}, c_{\max})$
        \State $h_i \sim \mathcal{U}(h_{\min}, h_{\max})$
        \State $k_i \sim \mathcal{U}(k_{\min}, k_{\max})$
        \State $g_i \sim \mathcal{U}(g_{\min}, g_{\max})$
        \State $P^0_{i} \leftarrow (\theta_i, \phi_i, a_i, c_i, h_i, k_i, g_i){}$
    \EndFor
    \For{$e=1,\dots, E$ epochs}
        \For{$t=1, \dots, T$ steps per epoch}
            \For{$i=1, \dots, M$ in population (\textbf{in parallel})}
                \State Collect exp. from env with $k_i$ and add to $\mathcal{D}$
            \EndFor
            \For{$i=1, \dots, M$ in population (\textbf{in parallel})}
                \For{$c=1, \dots, c_i$}
                    \State $\gamma_i = 1 - e^{g_i}$
                    \State \texttt{critic\_update}($\phi_i$, $\gamma_i$, $\mathcal{D}$)
                    \Comment{(Eq \ref{self-regularized})}
                \EndFor
                \For{$a=1, \dots, a_i$}
                    \State $H_i = h_i(-|\mathcal{A}|)$
                    \State \texttt{actor\_update}($\theta_i$, $H_i$, $\mathcal{D}$) \Comment{(Eq \ref{actor})}
                \EndFor
            \EndFor
        \EndFor
        
        \For{$i=1, \dots, M$ in population (\textbf{in parallel})}
            \State Evaluate $P^e_{i}$ for fitness $f_i$ with persistence $k_i$
        \EndFor
        \State Sort population $P$ by $f_i$
        \State ``Bad'' members $\leftarrow$ bottom $20\%$ of $P$
        \State ``Elite'' members $\leftarrow$ top $20\%$ of $P$
        \State Randomly shuffle ``Bad'' and ``Elite''
        \For{$bad \in \text{``Bad''}$ and $elite \in \text{``Elite''}$ }
            \State{Copy $elite$'s parameters \& weights to $bad$}
            \State{Perturb $bad$'s $a_i, c_i, h_i, k_i, g_i$}
        \EndFor

    \EndFor
\end{algorithmic}
\end{algorithm}
\end{minipage} 
\end{figure}

\section{Experiments}
\label{experiments}
We consider the following baselines:
\begin{itemize}
    \item \textbf{SAC}. Soft Actor-Critic with tunable entropy and literature-standard hyperparameters; a table of these standard parameters is available in Appendix \ref{baseline_impl}.
    \item \textbf{Persistence-Aware SAC ($k$-SAC)}. Soft Actor-Critic with tunable entropy and literature-standard hyperparameters, but trained with varying action persistence. We evaluate the agent on a range of $k$ values and report the highest performance.
    \item \textbf{Self-Regularized SAC (SR-SAC)}. We incorporate the self-regularized critic update (Eq \ref{self-regularized}) into standard SAC\footnote{We note that this agent is not equivalent to GRAC because it does not include its additional tricks (e.g., CEM action improvement). We are simply adding the self-regularized critic update to SAC to eliminate target networks.}. The number of critic updates per training step is determined with the heuristic in \cite{shao2020grac} - we update on a given batch until the loss drops below $\beta\%$ of its initial value. All other hyperparameters are set to the literature standards.
    \item \textbf{Random Parameter SAC (Rand-SAC)}. Soft Actor-Critic with hyperparameters uniformly chosen from AAC's search space\footnote{Default SAC has hyperparameters that AAC does not, e.g. $\tau$. In these cases, the value is chosen from a range around the literature default.}. Each run generates a new set of random hyperparameters. This highlights the hyperparameter sensitivity of SAC and shows the range of performance achieved by naively picking reasonable values to approach each environment.
\end{itemize}
The total network parameters are kept comparable by adding the action persistence value to the input state whether or not this value is varied during training. For example, SAC runs as normal with a state vector that has an additional element that is fixed at $1$. Results are listed as the mean and standard deviation of $5$ random seeds. Rand-SAC has high variance by design - we compensate for the extra randomness with $15$ total trials. The mean return of Rand-SAC is not as interesting as the variance because sufficient samples represent the performance of the mean of our random parameter distributions.

\begin{figure*}[h!]
\centering
\includegraphics[width=\textwidth]{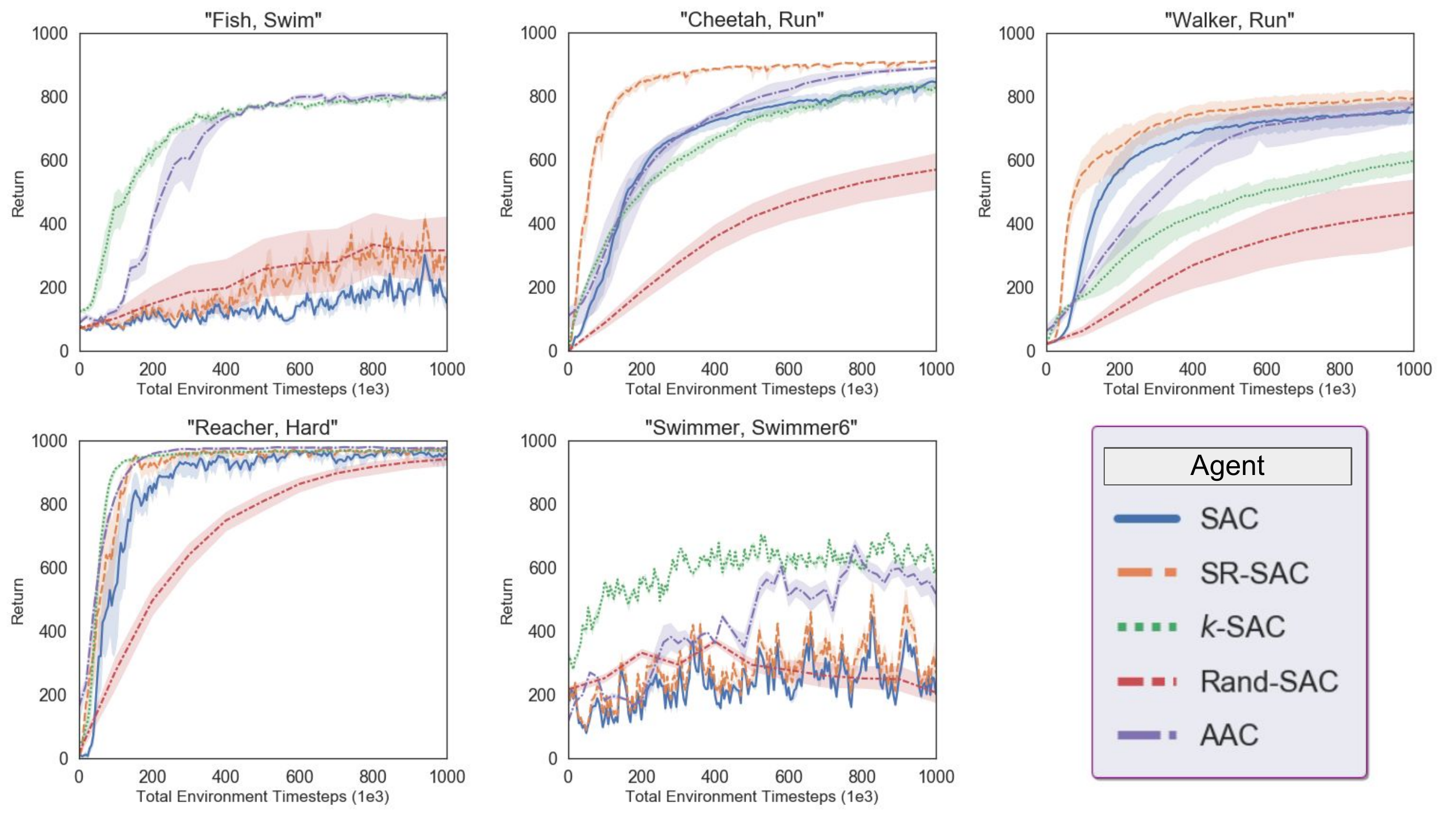}
\caption{\textbf{AAC on benchmark continuous control tasks.}  [Best viewed in color]}
\label{dmc_results}
\end{figure*}

The baselines are tested alongside AAC in five common tasks of varying difficulty from the DeepMind Control Suite \cite{tassa2018deepmind}. The results are shown in Figure \ref{dmc_results}. The randomized hyperparameters are consistently low-performance and high-variance, as expected. The standard SAC defaults are heavily tested on these tasks, so it is not surprising that they perform quite well. SR-SAC and $k$-SAC are special-purpose techniques designed to compensate for specific design choices in SAC. Their relative performance varies across each task and depends on whether the hyperparameter they address happens to be a significant factor. For example, $k=1$ is suboptimal in ``Fish, Swim", so $k$-SAC performs well. On the other hand, the critic learning schedule in default SAC is too conservative for ``Cheetah, Run", so SR-SAC offers a large improvement. AAC can adapt both of these parameters and discover the correct settings on a task-by-task basis; it matches the performance of the highest-performing baseline, although it may take more samples to sort out the optimal settings. Note that one reason for AAC's slight drop in sample efficiency is the value of $a_{max}$ and $c_{max}$ used in our experiments. We are not allowing our algorithm to fully compensate for the increase from $1$ environment sample per training step to AAC's distributed sampling. A member of the population that maxes out its actor and critic updates per step still cannot reach the replay ratio of SR-SAC. This choice was made because our implementation is synchronous, and allowing for a wide range in gradient update counts results in poor compute utilization. We discuss some workarounds for this in Section \ref{limitations}.

\begin{figure*}
\centering
\includegraphics[width=\textwidth]{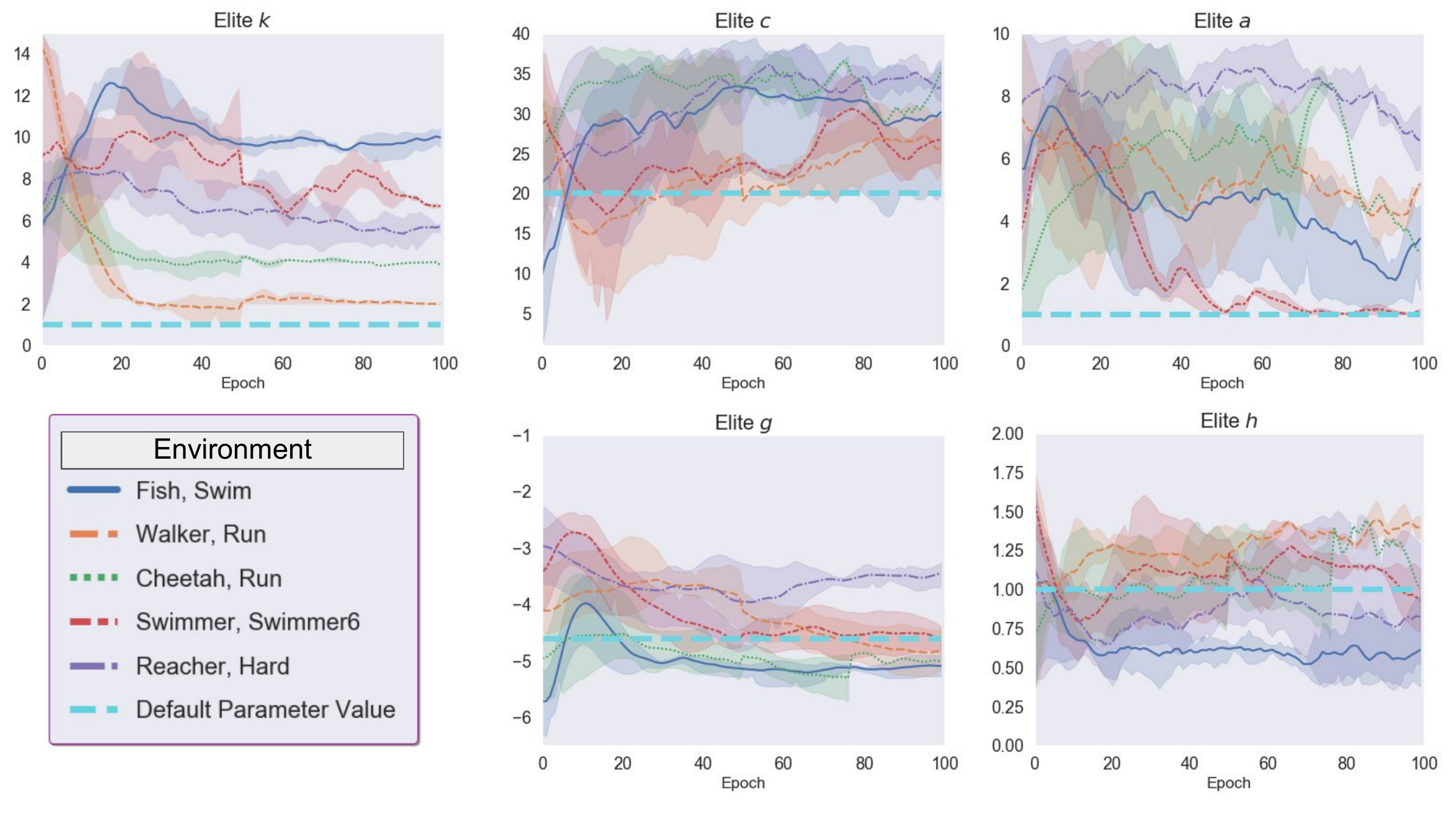}
\caption{\textbf{The hyperparameters learned by AAC.} The common default value is indicated with a horizontal blue line. $h$ and $g$ are substitute variables for $H$ and $\gamma$, respectively; see Sec \ref{method} for an explanation. [Best viewed in color]}
\label{dmc_hparam_results}
\end{figure*}

Figure \ref{dmc_hparam_results} shows the evolution of the highest-performing parameters over time. We plot the default parameter value as a light blue line for reference. AAC rediscovers the tuned default setting when it happens to be optimal for the task, e.g., $\gamma$ and $H$. Other parameters vary more across tasks, particularly $k$.

While it is helpful to know that AAC can find quality solutions to popular benchmarks, the real purpose of our algorithm is to simplify the use of actor-critic methods in less common domains. We put this to the test by evaluating AAC outside of simulated robotic locomotion. 

\begin{table*}[h!]
\centering
\resizebox{.9\textwidth}{!}{

\begin{tabular}{lcccccc}
\toprule[0.25ex]
& \textbf{Random Policy} & \textbf{Qin et al. \cite{qin2021neorl}} & \textbf{Rand-SAC} & \textbf{SAC} & \textbf{AAC} \\
\cmidrule(lr){1-6}
Setpoint 70 & $-322 \pm 57$ & $-180$ & $-399 \pm 99$ & $-216 \pm 16$ & $\mathbf{-175 \pm 3}$ \\
Setpoint 100 & $-439 \pm 129$ & - & $-432 \pm 147$ & $-314 \pm 56$ & $\mathbf{-257 \pm 43}$ \\
\bottomrule[0.25ex]
\end{tabular}
}
\caption{\textbf{Industrial Benchmark Results.} Total returns scaled by $1$e$-3$ for readability. The ``setpoint" parameter controls the difficulty of the environment and is bounded in $[0, 100]$. We add the setpoint 70 results from \cite{qin2021neorl} to verify our implementation.}
\label{ib_table}
\end{table*}

\begin{table*}[h!]
\centering
\resizebox{.9\textwidth}{!}{
\begin{tabular}{cccccc}
\toprule[0.25ex]
\multicolumn{1}{c}{\textbf{Random Policy}} & \multicolumn{1}{c}{\textbf{Rand-SAC}} & \multicolumn{1}{c}{\textbf{SAC}} & \multicolumn{1}{c}{\textbf{\begin{tabular}[c]{@{}c@{}}Hubbs et al. \cite{HubbsOR-Gym} \\ RL (PPO)\end{tabular}}} & \multicolumn{1}{c}{\textbf{\begin{tabular}[c]{@{}c@{}}Hubbs et al. \cite{HubbsOR-Gym} \\ Oracle\end{tabular}}} & \multicolumn{1}{c}{\textbf{AAC}} \\ \cmidrule(lr){1-6}
$8.8 \pm 81.1$ & $118 \pm 186$ & $342 \pm 11$ & $\mathbf{409.8 \pm 17.9}$ & $\mathit{542.7 \pm 29.9}$ & $\mathbf{415 \pm 1.5}$\\
\bottomrule[0.25ex]
\end{tabular}
}
\caption{\textbf{Inventory Management Results.} Rand-SAC, SAC and AAC collect $500,000$ steps of experience. The RL result from \cite{HubbsOR-Gym} uses Proximal Policy Optimization (PPO) \cite{schulman2017proximal}. The oracle method generates a theoretical upper bound by solving an optimization problem with information not available to the agent.}
\label{inv_table}
\end{table*}

\begin{table*}[h!]
\centering
\resizebox{.8\textwidth}{!}{
\begin{tabular}{l@{\hskip .5in}c@{\hskip .5in}c@{\hskip .5in}c@{\hskip .5in}c@{\hskip .5in}c}
\toprule[.25ex]
     & \textbf{Random Policy} & \textbf{SAC} & \textbf{Rand-SAC} & \textbf{AAC}    \\ \toprule[.25ex]
Mean & $-21,669$               & $-79$       & $-9,901$           & $\mathbf{2.68}$ \\
Std. & $21,032$                & $25$         & $11,008$           & $\mathbf{2.33}$ \\
Max  & $-1,015$                & $-23$        & $0.57$             & $\mathbf{6.83}$ \\ \bottomrule[.25ex]
\end{tabular}
}
\caption{\textbf{Newsvendor Results.} Total return (scaled by $1$e$-4$) in the environment over a $40$ day interval after $500,000$ environment steps. We also report the maximum score because asymmetric returns make the standard deviation a misleading estimate of the upper performance bound.}
\label{news_table}
\end{table*}

The Industrial Benchmark \cite{hein2017introduction} is a synthetic control task designed to imitate the challenges that arise in managing industrial systems. The agent controls three ``steering" variables and is rewarded for minimizing the cost and ``fatigue" associated with operating the system. The environment has stochastic and delayed rewards along with a partially observable state. We evaluate SAC, Rand-SAC, and AAC in two different situations of increasing difficulty. The results are displayed in Table \ref{ib_table}. AAC reduces the operating cost, and its effects are more noticeable at the greater difficulty.

Next we consider two inventory management problems (IMPs) proposed by \cite{HubbsOR-Gym} and \cite{balaji2019orl}. IMPs involve managing a supply chain to meet customer demand while balancing costs associated with ordering and carrying new materials\footnote{We refer the reader to the original references \cite{HubbsOR-Gym} \& \cite{balaji2019orl} for thorough descriptions of the environments.}. We assume no prior knowledge of the IMP and instead attempt to solve it using our automatic method. To demonstrate the benefits of an automated tuning system, we do not use an iterative development cycle\footnote{We make two changes to the DMC experiment settings before launch: $k_{max}$ is lowered to $5$ from $15$ because these environments have short time horizons of $40$ and $30$, and the epoch length is lowered to $500$ from $1000$ because we are testing after $500,000$ total timesteps.}; we ran AAC for five random seeds on each environment and report the initial results. The scores for the \texttt{InvManagement-v1} and \texttt{Newsvendor-v0} environments are listed in Tables \ref{inv_table} and \ref{news_table}, respectively. AAC outperforms our baselines and matches the performance of tuned RL results reported in \cite{HubbsOR-Gym}.

\begin{figure*}[h!]
    \centering
    \includegraphics[width=\textwidth]{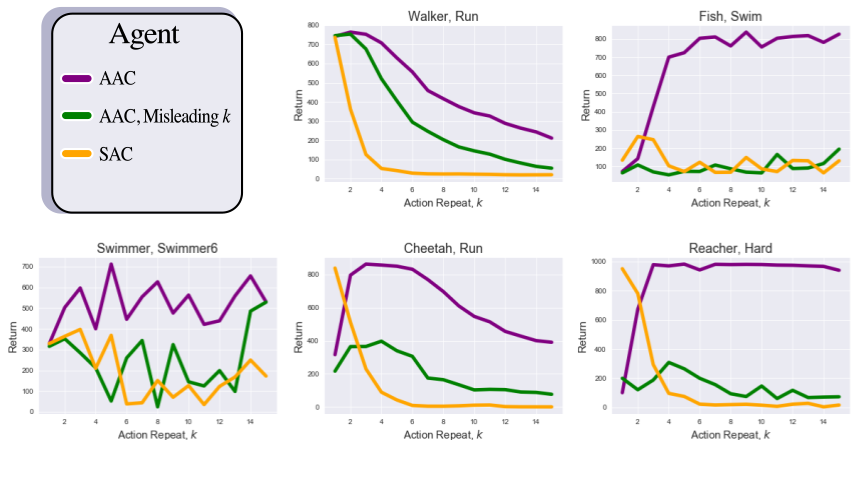}
    \caption{\textbf{Generalization of AAC across control frequencies}. Trained agents are evaluated across a range of control frequencies. We utilize AAC's ensemble of policies by returning the mean action across the population. AAC is more robust to changes in control frequency than SAC. In the ``Misleading $k$" experiments, the control frequency of the underlying environment is altered while the state representation of $k$ is fixed at $1$; the policy networks have learned to interpret the $k$ value to improve performance at sub-optimal control frequencies.}
    \label{generalization_fig}
\end{figure*}

Finally, we demonstrate the practical advantages of AAC's population of diverse and persistence-aware policies. Results on the DeepMind Control Suite environments are shown in Figure \ref{generalization_fig}. Ensembling the AAC population of actor networks greatly improves performance at sub-optimal control frequencies. Further experiments verify that the state representation of $k$ is correctly used to adapt the policy to changes in action persistence.

\section{Related Work}
This work contributes to the AutoRL literature of online hyperparameter tuning in Deep RL. \cite{hessel2019inductive} discusses algorithms' reliance on the inductive biases introduced by popular benchmarks and demonstrates that adaptive methods can match and exceed the performance of well-tuned baselines. OMPAC \cite{elfwing2017online} uses a genetic algorithm to select the policy's softmax temperature and TD($\lambda$) parameters in discrete environments such as Atari and Tetris. HOOF \cite{paul2019fast} generates a population of policy gradient updates with different loss function parameters and selects the best combination to continue training with weighted importance sampling. Agent57 \cite{badia2020agent57} uses hyperparameter selection by a multi-armed bandit to improve exploration and surpass human performance on the Atari benchmark. STAX \cite{zahavy2021selftuning} uses meta-gradients to tune the differentiable hyperparameters of the IMPALA \cite{espeholt2018impala} algorithm.

The two most similar works to our own are OHT-ES \cite{tang2020online} and SEARL \cite{franke2021sampleefficient}. Both are PBT-inspired hyperparameter tuning methods for off-policy RL with a shared replay buffer. OHT-ES adapts the learning rates and discount factor of TD3 while SEARL primarily adjusts the architecture of TD3's networks. Our method eliminates more RL hyperparameters using a MaxEnt framework, self-regularized critic update, dynamic learning schedule, and action repetition. However, we do not attempt to tune optimization-related hyperparameters like learning rate and network size. These methods are quite compatible, and it would be interesting to investigate extensions that combine our core agent with, e.g., the network search and tournament selection of SEARL. We do not compare against these works here because the differences in the core RL agent optimized and the parameters considered make the comparison unmeaningful. The action repetition in our method also complicates comparisons based on total timesteps. AAC prioritizes reducing hyperparameters and easing the development process, which is why our comparisons focus on variations of the RL agent we are optimizing.

\section{Limitations and Future Directions}
\label{limitations}
While our method successfully reduces RL-specific hyperparameters and design heuristics, we have not fully realized at least two of its promising advantages. First, distributed and diverse experience collection has the potential to increase exploration in sparse-reward environments. Diversity could be further improved by introducing exploration techniques from both the RL and evolutionary computation literature. We could motivate the exploration of individual actors by incorporating intrinsic rewards \cite{burda2018exploration}, and improve the parameter and behavioral diversity of the population as a whole with ideas from Novelty Search \cite{lehman2011abandoning} \cite{conti2017improving}.

We have also opted to avoid the meta-optimization of network-related hyperparameters such as model architecture and learning rate. The automatic discovery of optimal network architectures is an active area of research in the broader field of AutoML - see \cite{He_2021} for a survey. Many of these approaches could be added to our evolutionary algorithm with the help of an effective indirect encoding for model architecture and safe mutation operations. This is likely to increase the number of evolutionary epochs required to converge on a solution. However, there is plenty of evidence that network design can significantly increase the performance of actor-critic methods \cite{henderson2019deep}.

There is also room for improvement in terms of runtime and scalability. The synchronous implementation (see Appendix \ref{implementation}) used in our experiments limits our ability to adapt time-consuming parameters like the number of actor and critic gradient steps. The original PBT work \cite{jaderberg2017population} used an asynchronous framework where elite population members were checkpointed during training and could be read from disk when replacing low-performance members. A similar system could be adapted for the AAC algorithm. This would likely lead to a drop in sample efficiency but may open up the opportunity to scale the method to large clusters and search over more hyperparameters - especially network architectures.

\section{Conclusion}
This work has presented an automatic framework for online hyperparameter optimization in off-policy actor-critic algorithms. We have shown that our adaptive method can exceed the performance of tuned baselines in common benchmark tasks. However, the true promise of AutoRL methods lies in their ability to automate the process of engineering RL solutions to new domains. We demonstrated our algorithm's ability to succeed in less-studied industrial and operations research environments and are hopeful that this line of work will help enable the adoption of RL to a broader range of real-world problems. 

\printbibliography
\appendix

\section{RL Notation}
\label{mdp-notation}

\begin{itemize}
    \item $(s, a, r, s', d)$. One transition of experience from the environment. Consists of a state $s$, the action selected by the behavior policy $a$, the reward returned by the environment $r$ along with the next state $s'$ and boolean $d$ indicating the end of an episode.
    \item $\mathcal{A}$ is the action space. $|\mathcal{A}|$ refers to the dimension of the action space, or the number of elements in each action vector.
    \item $\gamma$. The discount factor that determines the agent's focus on long-term rewards. Gamma values approaching $1.0$ place encourage long-horizon planning while smaller values prioritize greedy behavior. The discounted expected return is defined:
    \begin{align}
        G(t) = \sum_{i=t}^{\infty}\gamma^ir_i
    \end{align}
    \item $\pi$. The policy function mapping states to a distribution over actions.
    \item $Q_{\pi}(s, a)$. The state-action value function, representing the expected discounted returns starting in state $s$, taking action $a$ and following policy $\pi$ thereafter.
    \item $V(s)$. The value function, representing the expected discounted returns starting in state $s$ and following policy $\pi$.
\end{itemize}

\section{Addressing Control Frequency with Persistence-Aware Actor-Critics}
\label{sec:persistence_critics}

As discussed in Sec \ref{design_choices}, the default control frequency of many environments is an arbitrary choice that can hinder optimization and exploration. When this is confronted in the literature, it is typically solved by making the action persistence a learned output of the actor or an additional set of discrete actions. However, the latter approach does not scale well with action size or maximum action repetition; providing a wide range of $k$ values for the Atari domain, for example, would require dozens of additional actions, greatly increasing the complexity of exploration. Instead, we typically pick one or two higher $k$ values above the single-step default. This replaces the control frequency hyperparameter with several new hyperparameters that likely require a grid search. Making $k$ a direct output of the actor network is a better solution for continuous domains, but it comes with implementation challenges of its own. For one, the discount $\gamma$ needs to be factored into the $k$ action repetitions - otherwise, the agent will favor high persistence values because they allow for the undiscounted accumulation of rewards. This can make it difficult to adjust $\gamma$ over the course of off-policy training. Furthermore, dynamic action repetition begins to trespass on the territory of hierarchical RL; a $k$-repetition policy can be formalized as a $k$-step \textit{option} of a MDP. A policy that operates on an unpredictable timescale may complicate integration with higher-level policies.

We consider an alternative solution in which the action persistence value $k$ is an element of the state vector. The actor and critic networks are able to interpret the current persistence and adapt accordingly. We can then vary the persistence throughout training and evaluate the same actor network on multiple $k$ values in order to determine the appropriate setting at test time. We make another modification and have the environment return an array of rewards that correspond to each of the $k$ possible timesteps. This lets us adjust the value of $\gamma$ and recompute accurate TD targets during the critic update by discounting each element of the array and discounting the subsequent $Q$ value by an additional timestep. 

\begin{figure*}[!htb]
\minipage{0.32\textwidth}
  \includegraphics[width=\linewidth]{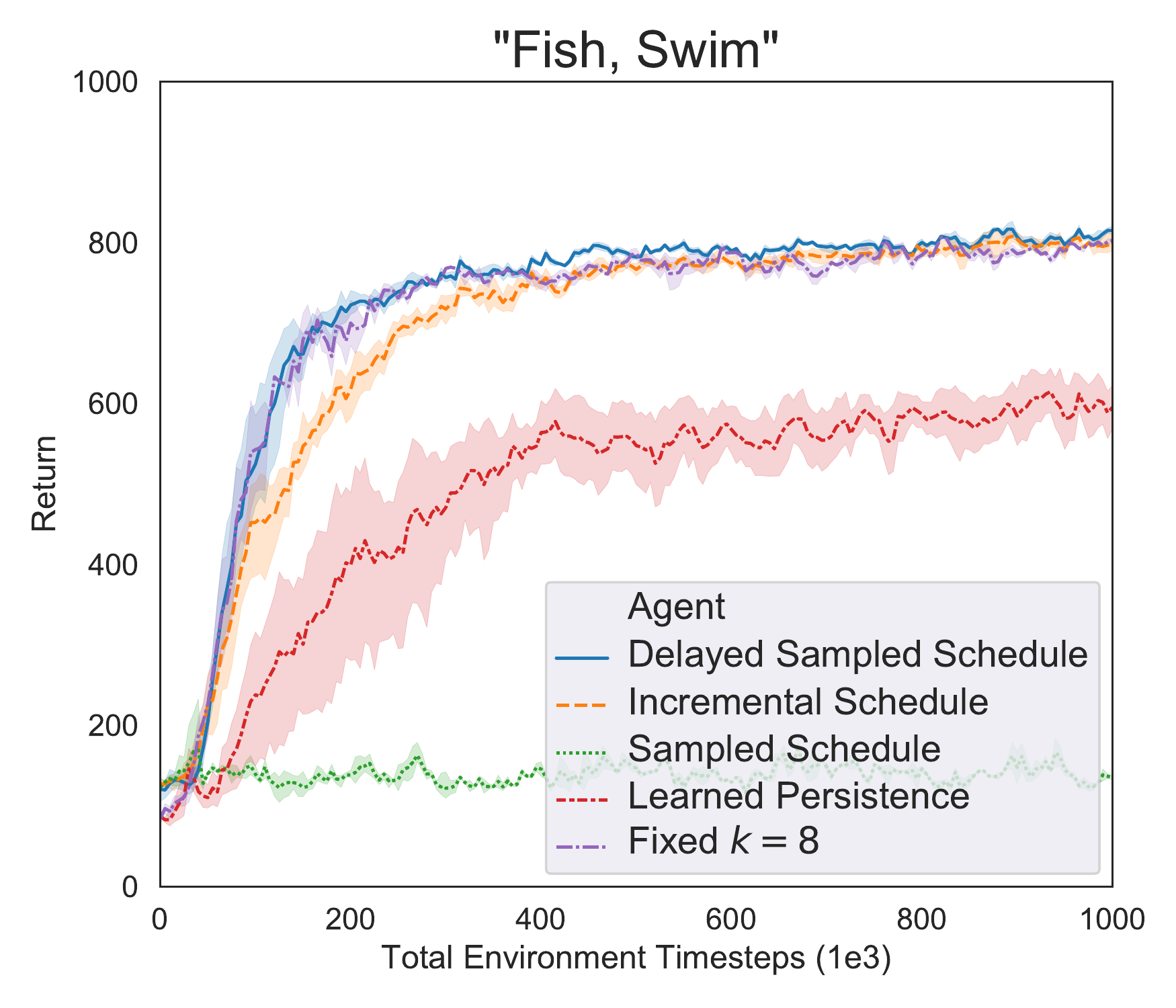}
\endminipage\hfill
\minipage{0.32\textwidth}
  \includegraphics[width=\linewidth]{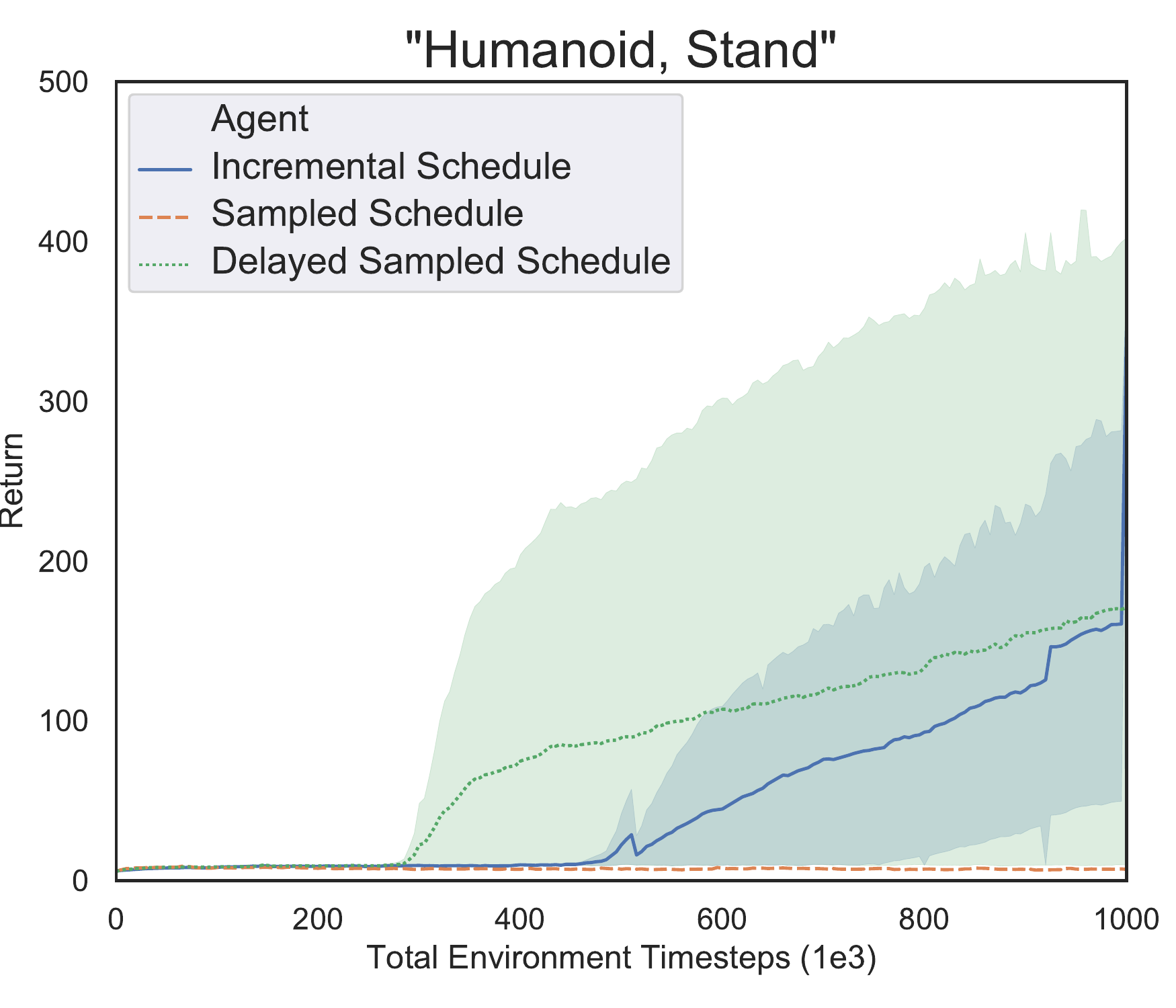}
\endminipage\hfill
\minipage{0.32\textwidth}%
  \includegraphics[width=\linewidth]{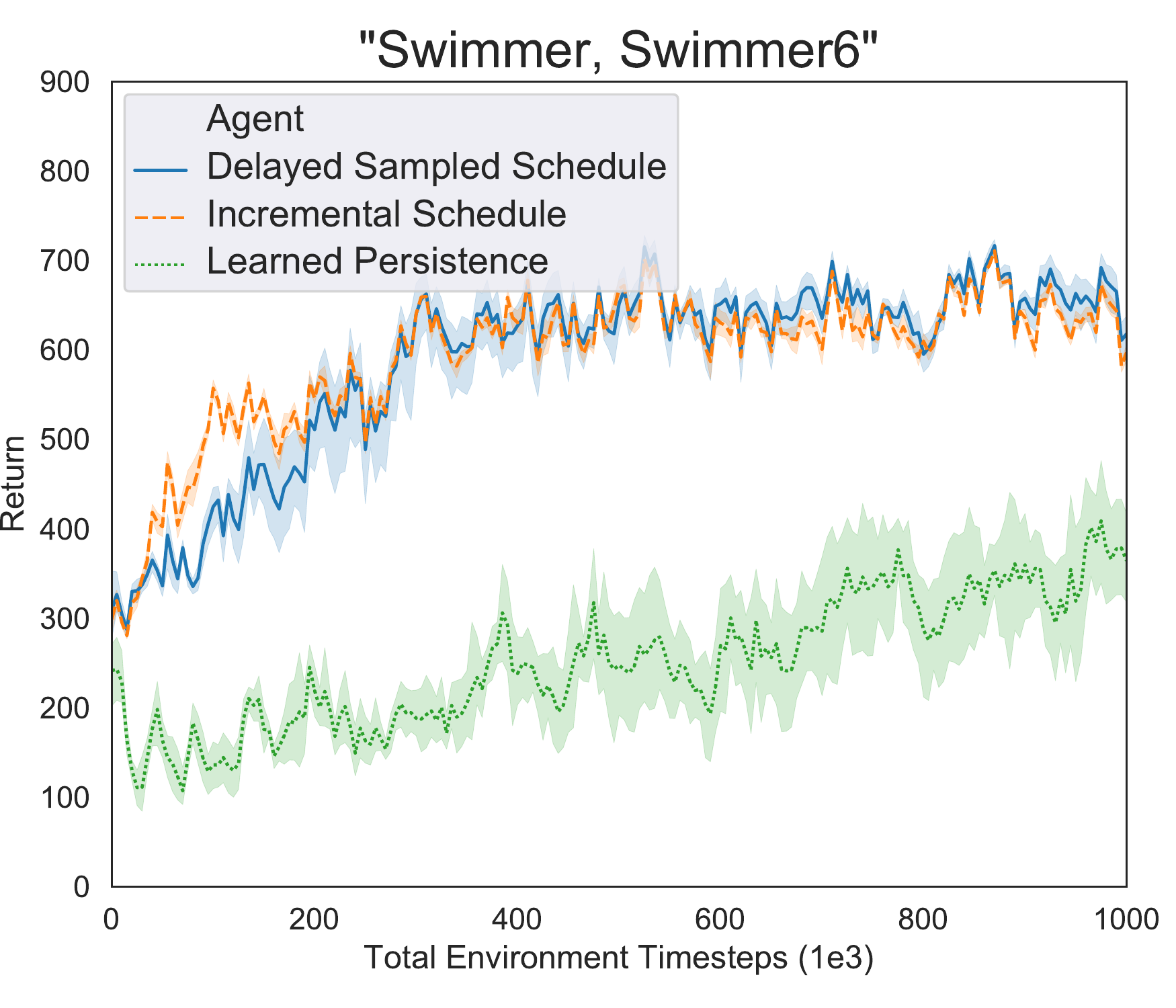}
\endminipage
\caption{\textbf{Action Persistence Experiments.}}
\label{persistence-demo}
\end{figure*}

We demonstrate this technique in three tasks from the DeepMind Control Suite. ``Learned Persistence" adds $k$ as an additional element of the action space. ``Incremental Schedule" methods use the naive approach of iterating through all reasonable $k$ values during training. ``Sampled Schedule" uses Thompson sampling over the returns at the previous evaluation period to pick $k$ for the next training phase. ``Delayed Sampled Schedule" begins by iterating through all $k$ values as a crude exploration mechanism before sampling the setting later in training to avoid wasting time on clearly sub-optimal settings. All methods are evaluated on all $k$ values and the highest return is reported. Results are shown in Figure \ref{persistence-demo}.

A side effect of this approach is that it allows the agent to generalize across control frequencies and therefore adapt to changes that may occur during deployment. If the training phase determines that the optimal action persistence is $5$, for example, then a sensor malfunction or slowdown that cuts our control frequency in half can be compensated for by setting the persistence element of the agent's state vector to $10$. This is still likely to decrease performance, but in our case the agent has seen values of $10$ during training and is more capable of generalizing to the new situation. This effect is demonstrated by experiments in Figure \ref{generalization_fig}.

\section{AAC Algorithm Details}
\subsection{Implementation Details}
\label{implementation}

We initialize a population of $20$ members with hyperparameters chosen uniformly from the range provided in Table \ref{search-range-table}. This range was determined by simple intuition about the range of reasonable parameter settings that we might otherwise grid search over. The only unintuitive choice is the difference between $c_{\max}$ and $a_{\max}$, which is based on the need for the self-regularized TD update (Eq \ref{self-regularized}) to perform many more critic updates than is standard. The $i$th member of the population has parameters ($\theta_i, \phi_i, a_i, c_i, h_i, k_i, g_i$). We use the standard clipped-double-Q-trick \cite{fujimoto2018addressing} such that each agent actually has two critic networks $\phi_{1_i}$ and $\phi_{2_i}$.

We collect $10,000$ random environment samples split evenly among the full range of $k$ values to initialize the replay buffer. Each agent is trained in parallel, adding experience from the environment with an action persistence of $k_i$ to a collective replay buffer that holds $2,000,000$ samples before overwriting the oldest experience\footnote{We train the agent for a maximum of $2,000,000$ total samples, and only displayed the results after $1,000,000$, so experience is never overwritten in practice.}. Therefore each iteration of the AAC algorithm collects $20$ new transitions. 

During each training step, agent $i$ updates its critic networks $c_i$ times and its actor network $a_i$ times. Each of these training steps samples a fresh batch of experience from the buffer with a batch size of $512$ for DMC experiments and $128$ otherwise. This is a slight divergence from the self-regularized heuristic of GRAC where the same batch is optimized repeatedly. We utilize an environment wrapper that returns an array of rewards $\mathbf{\hat{r}}$ representing the reward at each timestep up to the maximum possible $k$ value. If $k_i < k_{\max}$ the extraneous entries are set to $0$. When computing temporal difference targets, we multiply the $j$th entry of $\mathbf{\hat{r}}$ by $\gamma^j$ and sum the resulting array to get the $r$ term in Eq \ref{critic}. We then multiply the $Q_{\phi_i}(s', a')$ term by $\gamma^{k_i + 1}$ to keep the time horizon consistent across different action repetitions. We also override terminal signals for episodes that end as a result of reaching the max step limit\footnote{This implementation detail is known as ``infinite bootstrapping" and analyzed in \cite{pardo2018time}.}. The $\alpha$ entropy coefficient is updated using the gradient descent technique from SAC \cite{haarnoja2019soft} with target entropy $h_i( -|\mathcal{A}|)$ - this takes place inside the actor update function, meaning $\alpha$ is updated $a_i$ times per training step. 

\begin{table}[h]
\centering
\begin{tabular}{@{}clll@{}}
\toprule
\textbf{Param} & \textbf{Min}  & \textbf{Max}  & $\mathbf{\delta}$ \\ \midrule
$a$   & 1    & 10   & 2     \\
$c$   & 1    & 40   & 5     \\
$h$   & 0.25 & 1.75 & 0.25  \\
$k$   & 1    & 15 (DMC, IB), 5 (OR-Gym)   & 2     \\
$g$   & -6.5 & -1   & 0.5   \\ \bottomrule
\end{tabular}
\vspace{2mm}
\caption{\textbf{Parameter search ranges.} See explanation of each in Sec \ref{method}.}
\label{search-range-table}
\end{table}

Each agent continues to train independently for one evolutionary epoch. The length of each epoch is a new hyperparameter of our method. We set this length to be $1,000$ steps, which corresponds to $1$ episode of training in the DeepMind Control Suite environments. This choice is arbitrary and likely to be sub-optimally short but achieves good results and reduces training time. At the end of each epoch, the population is evaluated. The fitness of each member in the population is set to the mean return across these evaluations. We sort the population by fitness and separate the best $20\%$ and worst $20\%$ of agents. The choice of $20\%$ as a threshold is so arbitrary that attempting to tune it would be against the spirit of our ``automatic" approach - this value is simply copied from Population Based Training \cite{jaderberg2017population} and was never changed during our research process. These ``bad" and ``elite" groups are randomly paired for evolutionary updates. Each pair copies the values ($\theta_i, \phi_i, a_i, c_i, g_i, h_i, k_i$) along with the current $\alpha$ and Adam optimizer \cite{kingma2017adam} settings from the elite agent to the bad agent. The hyperparameters $a_i, c_i, g_i, h_i, k_i$ are then altered by adding a perturbation sampled uniformly from the range $[-\delta, \delta]$\footnote{The perturbations for integer hyperparameters like $a, c$ and $k$ are sampled uniformly from the integers in the range $(-\delta, \delta)$.}. The values of $\delta$ for each parameter are also listed in Table \ref{search-range-table}. During our development process, these values were determined by guessing an appropriate range and then given a slight boost to generate a satisfactory shift in the parameter distributions over time.

Our synchronous implementation runs at around $2$ iterations per second ($T$ in Algorithm \ref{algo}), leading to a training time of $7$ hours for the main DeepMind Control Suite experiments. The population is split across $2$ GPUs. At 2 GPUs and 7 hours per trial for at least 3 trials in 9 environments, the AAC results in this paper consume approximately $378$ GPU hours. The SAC, Rand-SAC and $k$-SAC baselines train in roughly $3$ hours on a single GPU. With 3 algorithms in 9 environments running for $5$ random seeds\footnote{We use $15$ random seeds for Rand-SAC, which has much higher variance by design.}, the baselines consume approximately $216$ GPU hours.

\section{Baseline Implementation Details}
\label{baseline_impl}
Our default hyperparameters for SAC and SR-SAC are listed in Table \ref{literature-hparams}. These settings are chosen based on \cite{pytorch_sac}, \cite{fujimoto2018addressing}, \cite{haarnoja2019soft}, and other publicly available implementations.

\begin{table}[h]
\centering
\begin{tabular}{ll}
\toprule
\textbf{Param}                & \textbf{Value}                    \\ \toprule
Batch Size           & $128$, $512$ (DMC)       \\
$\tau$               & $.005$                   \\
actor lr             & $3e$-$4$                   \\
critic lr            & $3e$-$4$                   \\
$\alpha$ lr             & $1e$-$4$                   \\
$\gamma$             & $0.99$                   \\
Warmup Steps         & $1000$                   \\
Target Delay         & $2$                      \\
Critic Updates Per Step & $1$                 \\
Actor Updates Per Step & $1$                 \\
SR $\beta$ Init      & $70$                     \\
SR $\beta$ Final     & $90$                     \\
$H$                  & $-|\mathcal{A}|$         \\
Architecture         & $256$, ReLU, $256$, ReLU \\
Action Log Std Range & $(-10, 2)$              \\
\bottomrule
\end{tabular}
\vspace{2mm}
\caption{\textbf{Standard hyperparameters for SAC, SR-SAC, and $k$-SAC used in our experiments.}}
\label{literature-hparams}
\end{table}

\end{document}